\documentclass[letterpaper, 10 pt, conference]{ieeeconf}  %

\IEEEoverridecommandlockouts                              %

\usepackage{times}
\usepackage{epsfig}
\usepackage{graphicx}
\usepackage{amsmath}
\usepackage{amssymb}
\usepackage{balance}
\usepackage[percent]{overpic}
\usepackage{pifont}%
\usepackage{wasysym}

\usepackage[utf8]{inputenc} %
\usepackage[T1]{fontenc}    %
\usepackage{xcolor}         %
\definecolor{mypink1}{rgb}{0.78, 0.21, 0.13}
\usepackage[colorlinks=true]{hyperref}       %
\hypersetup{
    colorlinks=true,
    linkcolor=mypink1,
    filecolor=magenta,      
    }
\usepackage{url}            %
\usepackage{amsfonts}       %
\usepackage{nicefrac}       %
\usepackage{microtype}      %
\usepackage{tabularx}
\usepackage{booktabs}

\usepackage{graphicx}
\usepackage[export]{adjustbox}
\usepackage{xcolor}
\usepackage{overpic}
\usepackage{multirow}

\usepackage{times}
\usepackage{epsfig}
\usepackage{graphicx}
\usepackage{amsmath}
\usepackage{amssymb}
\usepackage{balance}
\usepackage[percent]{overpic}
\usepackage{pifont}%
\usepackage{wasysym}
\usepackage{booktabs}

\title{\LARGE \bf
Mask3D: Mask Transformer for 3D Semantic Instance Segmentation 
}

\usepackage{booktabs}
\usepackage{balance}
\usepackage{multirow}
\usepackage{cite}

\usepackage[font=small,labelfont=bf]{caption}
\usepackage{color}
\usepackage{xcolor}
\usepackage{tikz}
\usepackage{pgfplots}
\pgfplotsset{compat=1.17}
\usepackage{tabularx}
\usepackage{amssymb}%
\usepackage{pifont}%

\usepackage{amsmath}

\usepackage{MnSymbol}%
\usepackage[super]{nth}

\usepackage{colortbl}
\usepackage{subfigure}
\usepackage{scalefnt}
\usepackage{tabu}
\usepackage{wrapfig}
\usepackage{wasysym}%
\usepackage{xspace}

\newcommand{\refsec}[1]{Sec.\,\ref{sec:#1}}

\newcommand{\refeq}[1]{Eq.\,\ref{eq:#1}}
\newcommand{\reffig}[1]{Fig.\,\ref{fig:#1}}
\newcommand{\reftab}[1]{Tab.\,\ref{tab:#1}}

\def\eg{\emph{e.g.}\@\xspace} 
\def\ie{\emph{i.e.}\@\xspace} 
\def\cf{\emph{c.f.}\@\xspace} 
 
\def\etal{\emph{et al.}\@\xspace}

\newcommand{\circlenum}[1]{{\textcircled{\scriptsize{#1}}}}
\newcommand{\parag}[1]{\vskip2pt \noindent \textbf{#1}}

\definecolor{darkgreen}{RGB}{0,255,0}
\definecolor{linkgreen}{RGB}{52,130,48}
\definecolor{m_green}{RGB}{233, 254, 187}
\definecolor{m_orange}{RGB}{255, 212, 121}
\definecolor{m_red}{RGB}{255, 190, 188}
\definecolor{m_violet}{RGB}{215, 131, 255}
\definecolor{m_blue}{RGB}{186, 234, 255}
\newcommand{\colorsquare}[1]{{\color{#1}$\blacksquare$}\hspace{-6.4pt}$\square$}

\newcommand{\cmark}{\ding{51}}%
\newcommand{\xmark}{\ding{55}}%

\newcommand{\name}{Mask3D}

\newif\ifmynotes
\mynotestrue 
\definecolor{notetext}{rgb}{0.7,0,0}

\newcolumntype{Y}{>{\centering\arraybackslash}X}
\newcolumntype{Z}{>{\raggedleft\arraybackslash}X}

\author{
Jonas Schult$^{1}$,
Francis Engelmann$^{2,\,3}$,
Alexander Hermans$^{1}$,
Or Litany$^{4}$,
Siyu Tang$^{2}$,
Bastian Leibe$^{1}$%
\thanks{%
$^{1}$ Computer Vision Group, RWTH Aachen University, Germany.}%
\thanks{$^{2}$ Computer Vision and Learning Group, ETH Z\"{u}rich, Switzerland.}%
\thanks{$^{3}$ ETH AI Center, Z\"{u}rich, Switzerland.}%
\thanks{$^{4}$ NVIDIA, Santa Clara, USA}%
}

\begin{document}

\maketitle
    \thispagestyle{empty}
\pagestyle{empty}

\begin{abstract}
Modern 3D semantic instance segmentation approaches predominantly rely on specialized voting mechanisms followed by carefully designed geometric clustering techniques.
Building on the successes of recent Transformer-based methods for object detection and image segmentation,
we propose the first Transformer-based approach for 3D semantic instance segmentation.
We show that we can leverage generic Transformer building blocks to
directly predict instance masks from 3D point clouds.
In our model -- called \name{} -- each object instance is represented as an \emph{instance query}. 
Using Transformer decoders, the instance queries are learned by iteratively attending to point cloud features at multiple scales. 
Combined with point features, the instance queries directly yield all instance masks in parallel.
\name{} has several advantages over current state-of-the-art approaches, since it neither relies on (1) voting schemes which require hand-selected geometric properties (such as centers) nor (2) geometric grouping mechanisms requiring manually-tuned hyper-parameters (\eg radii) and (3) enables a loss that directly optimizes instance masks.
\name{} sets a new state-of-the-art on ScanNet test (+\,6.2\,mAP), {S3DIS 6-fold  ({+\,10.1\,}\,mAP)}, {STPLS3D ({+\,11.2}\,mAP)} and {ScanNet200 test (+\,12.4\,}mAP). %
\end{abstract}

\section{Introduction}
This work addresses the task of semantic instance segmentation of 3D scenes.
That is, given a 3D point cloud, the desired output is a set of object instances represented as binary foreground masks (over all input points) with their corresponding semantic labels (\eg \emph{`chair'}, \emph{`table'}, \emph{`window'}).

Instance segmentation resides at the intersection of two problems:
semantic segmentation and object detection.
Therefore methods have opted to either first learn semantic point features followed by grouping them into separate instances (bottom-up) or detecting object instances followed by refining their semantic mask (top-down).
Bottom-up approaches (ASIS\cite{Wang19CVPR}, SGPN\cite{Wang18CVPR}, 3D-BEVIS\cite{Elich19GCPR}) employ contrastive learning, mapping points to a high-dimensional feature space where features of the same instance are close together, and far apart otherwise.
Top-down methods (3D-SIS\cite{Hou19CVPR}, 3D-BoNet\cite{Yang19NIPS}) use an approach akin to Mask R-CNN\cite{He17ICCV}:
First detect instances as bounding boxes and then perform mask segmentation on each box individually.
While 3D-SIS\cite{Hou19CVPR} relies on predefined anchor boxes\cite{He17ICCV},
3D-BoNet\cite{Yang19NIPS} proposes an interesting variation that predicts bounding boxes from a global scene descriptor and optimizes an association loss based on
{bipartite matching}\cite{Kuhn55hungarian}.
A major step forward was sparked by powerful feature backbones \cite{Thomas19ICCV,Wang17TOG,Graham17ARXIV} such as sparse convolutional networks \cite{Choy19CVPR, Graham18CVPR} that improve over
existing PointNets\cite{Qi17CVPR,Qi17NeurIPS} and dense 3D CNNs\cite{Maturana15IROS, Wu15CVPR,Qi16CVPR}.
Well established 2D CNN architectures \cite{He2016CVPR, Ronneberger2015MICCAI} can now easily be  adapted to sparse 3D data.
These models can process large-scale 3D scenes in one pass, which is necessary to capture global scene context at multiple scales.
As a result, bottom-up approaches which benefit from strong features (MTML\cite{Lahoud19ICCV}, MASC\cite{Liu19ARXIV}) experienced another performance boost.
Soon after, inspired by Hough voting approaches \cite{Hough59,Leibe08IJCV}, VoteNet\cite{Qi19ICCV} proposed \emph{center-voting} for 3D object detection.
Instead of mapping points to an abstract high-dimensional feature space (as in bottom-up approaches), points now vote for their object center  {-- votes from the same object are then closer to each other which enables geometric grouping into instance masks.}
This idea quickly influenced the 3D instance segmentation field, and by now,
the vast majority of current state-of-the-art 3D instance segmentation methods \cite{Chen21ICCV, Engelmann20CVPR, Han20CVPR, Jiang20CVPR, Vu22CVPR}
make use of both object center-voting and sparse feature backbones.

\begin{figure}
    \begin{overpic}[abs,unit=1mm,scale=.25,width=1.0\linewidth]{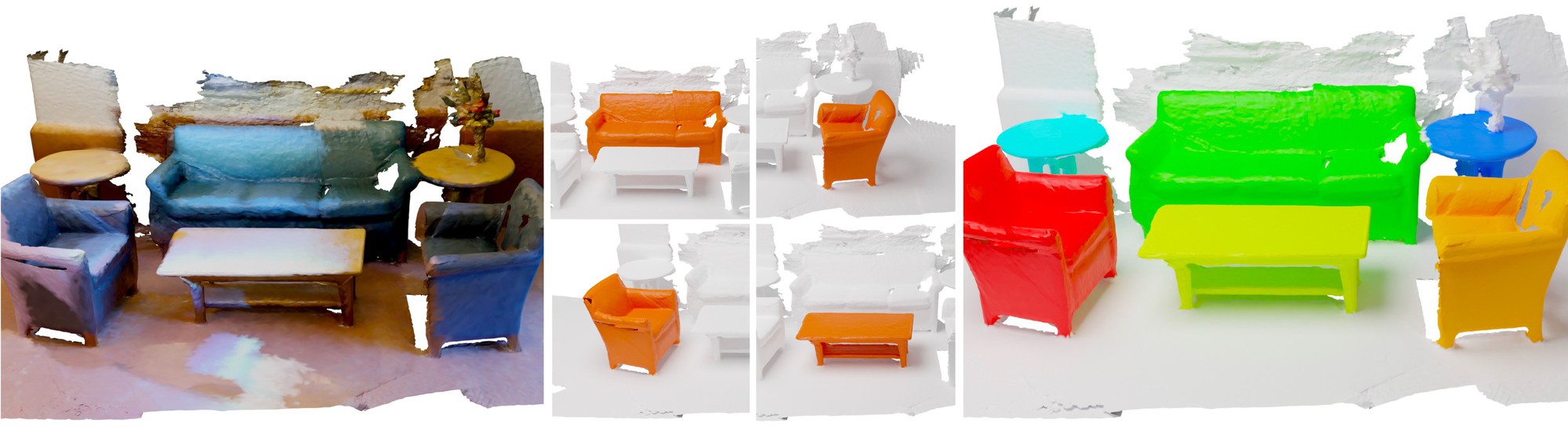}
    \put(5.3,0){\footnotesize \colorbox{white}{Input 3D Scene}}
    \put(29.7,0){\footnotesize  \colorbox{white}{Instance Heatmaps}}
    \put(55.9,0){\footnotesize \colorbox{white}{3D Semantic Instances}}
    \vspace{-25px}
    \end{overpic}
    \caption{\textbf{\name{}.}
    We train an end-to-end model for 3D semantic instance segmentation on point clouds.
    Given an input 3D point cloud \emph{(left)}, our Transformer-based model uses an attention mechanism to produce instance heatmaps {across all points} \emph{(center)} and directly predicts all semantic object instances in parallel \emph{(right)}.
    }
    \label{fig:teaser}
\end{figure}

Although 3D instance segmentation has made impressive progress, current approaches have several major problems:
typical state-of-the-art models are based on manually-tuned components, such as
voting mechanisms that predict hand-selected geometric properties (\eg{},
centers\cite{Jiang20CVPR}, bounding boxes\cite{Chibane2022arxiv}, occupancy\cite{Han20CVPR}),
and heuristics for clustering the votes (\eg{}, dual-set grouping \cite{Jiang20CVPR}, proposal aggregation\cite{Engelmann20CVPR}, set aggregation/filtering \cite{Chen21ICCV}).
Another limitation of these models is that they are not designed to directly predict instance masks.
Instead, masks are obtained by grouping votes, and the model is trained using proxy-losses on the votes. %
A more elegant alternative consists of directly predicting and supervising instance masks, such as 3D-BoNet\cite{Yang19NIPS} or DyCo3D\cite{He21CVPR}.
Recently, this idea gained popularity in 2D object detection (DETR\cite{Carion20ECCV}) and image segmentation (Mask2Former\cite{Cheng21NeurIPS,Cheng22CVPR}) 
but so far received less attention in 3D \cite{Yang19NIPS, He21CVPR, Misra21ICCV}.
At the same time, in 2D image processing, we observe a strong shift from ubiquitous CNN architectures \cite{Ronneberger2015MICCAI,renNIPS15fasterrcnn, He2016CVPR, He17ICCV} towards Transformer-based models \cite{dosovitskiy2020vit, liu2021Swin, Cheng21NeurIPS}.
\\
In 3D, the move towards Transformers is less pronounced with only
a few methods focusing on 3D object detection \cite{Misra21ICCV, Liu21ICCV, Pan21CVPR} or 3D semantic segmentation \cite{Lai22CVPR, Zhao21ICCV, Zhao21ICCV}
and no methods for 3D instance segmentation.
Overall, these approaches are still behind in terms of performance compared to current state-of-the-art methods \cite{Vu22CVPR, Nekrasov2021ThreeDV, Wang17TOG, Rukhovich21arxiv}.

In this work, we propose the first Transformer-based model for 3D semantic instance segmentation of large-scale scenes that sets new state-of-the-art scores over a wide range of datasets, and addresses the aforementioned problems on hand-crafted model designs.
The main challenge lies in directly predicting instance masks and their corresponding semantic labels.
To this end, our model
predicts \emph{instance queries} that encode semantic and geometric information of each instance in the scene.
Each instance query is then further decoded into a semantic class and an \emph{instance feature}.
The key idea (to directly generate masks) is to compute {similarity scores between individual instance features and all point features} in the point cloud\cite{He21CVPR, Chen21ICCV, Cheng21NeurIPS}.
This results in a heatmap over the point cloud, which (after normalization and thresholding) yields the final binary instance mask (\cf \reffig{teaser}).
Our model, called \name{}, builds on recent advances in both Transformers \cite{Cheng22CVPR, Misra21ICCV} and 3D deep learning \cite{Choy19CVPR, Graham17ARXIV, Wang17TOG}:
to compute strong point features, we leverage a sparse convolutional feature backbone \cite{Choy19CVPR} that efficiently processes full scenes and naturally provides multi-scale point features.
To generate instance queries, we rely on stacked Transformer decoders \cite{Cheng21NeurIPS, Cheng22CVPR} that iteratively attend to learned point features in a coarse-to-fine fashion  using non-parametric queries \cite{Misra21ICCV}.
Unlike voting-based methods, directly predicting and supervising masks causes some challenges during training:
before computing a mask loss,
we first have to establish correspondences between predicted and annotated masks.
A na\"ive solution would be to choose for each predicted mask the nearest ground truth mask \cite{He21CVPR}.
However, this does not guarantee an optimal matching and any unmatched annotated mask would not contribute to the loss.
Instead, we perform bipartite graph matching to obtain optimal associations between ground truth and predicted masks\cite{Yang19NIPS,Carion20ECCV}. %
We evaluate our model on four challenging 3D instance segmentation datasets, ScanNet\,v2\cite{Dai17CVPR}, ScanNet200\cite{rozenberszki22ECCV}, S3DIS\cite{Armeni16CVPR} and STPLS3D\cite{Chen22Arxiv}
and significantly outperform prior art, even surpassing architectures that are highly tuned towards specific datasets.
Our experimental study compares various query types,
different mask losses, and evaluates the number of queries as well as Transformer decoder steps. 

Our contributions are as follows:
\textbf{(1)} We propose the first competitive Transformer-based model for 3D semantic instance segmentation.
\textbf{(2)} Our model named \name{} builds on domain-agnostic components, avoiding center voting, non-maximum suppression, or grouping heuristics, 
 and overall requires less hand-tuning. %
\textbf{(3)} \name{} achieves state-of-the-art performance on ScanNet, ScanNet200, S3DIS and STPLS3D.
To reach that level of performance with a Transformer-based approach,
it is key to predict instance queries that encode the semantics and geometry of the scene and objects.

\section{Related Work}

\parag{3D Instance Segmentation.}
Numerous methods have been proposed for 3D instance semantic segmentation,
including bottom-up approaches \cite{Wang19CVPR, Wang18CVPR, Elich19GCPR, Lahoud19ICCV, Liu19ARXIV}, top-down approaches \cite{Hou19CVPR, Yang19NIPS, Yang19NIPS}, and more recently, voting-based approaches \cite{Chen21ICCV, Engelmann20CVPR, Han20CVPR, Jiang20CVPR, Vu22CVPR}.
MASC\cite{Liu19ARXIV} uses a multi-scale hierarchical feature backbone, similar to ours,
however, the multi-scale features are used to compute pairwise affinities followed by an offline clustering step.
Such backbones are also successfully employed in other fields \cite{Cheng22CVPR,Rukhovich21arxiv}.
Another influential work is DyCo3D\cite{He21CVPR}, which is among the few approaches that directly predict instance masks without a subsequent clustering step.
DyCo3D relies on \emph{dynamic convolutions}\cite{Jia16NIPS, Tian20ECCV} which is similar in spirit to our mask prediction mechanism.
However, it does not use optimal supervision assignment during training, resulting in subpar performance.
Optimal assignment of the supervision signal was first implemented by 3D-BoNet\cite{Yang19NIPS} using Hungarian matching.
Similar to ours, \cite{Yang19NIPS} directly predicts all instances in parallel. However, it uses only a single-scale scene descriptor which cannot encode object masks of diverse sizes.

\parag{Transformers.}
Initially proposed by Vaswani~\etal\cite{Vaswani17NeurIPS} for NLP,
Transformers have recently revolutionized the field of computer vision with successful models such as ViT\cite{dosovitskiy2020vit} for image classification, DETR\cite{Carion20ECCV} for 2D object detection,
or Mask2Former\cite{Cheng21NeurIPS,Cheng22CVPR} for 2D segmentation tasks.
The success of Transformers has been less prominent in the 3D point cloud domain though and recent Transformer-based methods focus on either 3D object detection \cite{Misra21ICCV, Liu21ICCV, Pan21CVPR} or 3D semantic segmentation \cite{Lai22CVPR, Zhao21ICCV,Park22CVPR}.
Most of these rely on specific attention modifications to deal with the quadratic complexity of the attention \cite{Pan21CVPR,Zhao21ICCV,Lai22CVPR,Park22CVPR}.
Liu \etal\cite{Liu21ICCV} use vanilla Transformer decoder, but only to refine object proposals, whereas Misra \etal\cite{Misra21ICCV} are the first to show how to apply a vanilla Transformer to point clouds, still relying on an initial learned downsampling stage though.
DyCo3D\cite{He21CVPR} also uses a Transformer, however at the bottleneck of the feature backbone to increase the receptive field size and is not related to our mechanism for 3D instance segmentation.
In this work, we show how a vanilla Transformer decoder can be applied to the task of 3D semantic instance segmentation {and achieve state-of-the-art performance.}

\begin{figure}[t!]
\vspace{5pt}
\includegraphics[width=\linewidth]{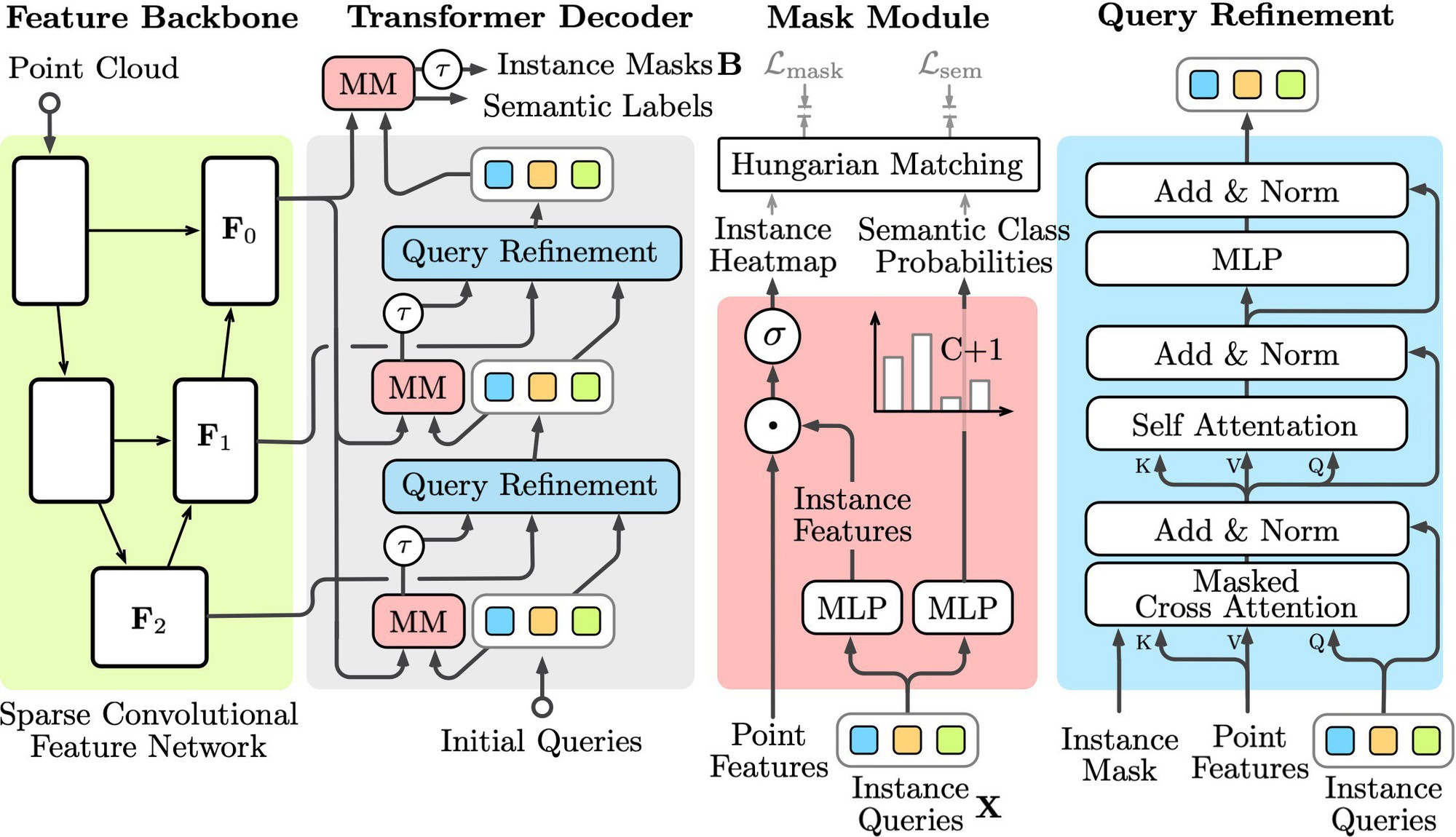}
\vspace{-15pt}
\caption{
\textbf{Illustration of the \name{} model.}
The feature backbone outputs multi-scale point features $\mathbf{F}$, while the Transformer decoder iteratively refines the instance queries $\mathbf{X}$.
Given point features and instance queries, the mask module predicts for each query a semantic class and an instance heatmap, which (after thresholding) results in a binary instance mask $\mathbf{B}$. 
\circlenum{$\tau$}
 applies a threshold of 0.5 and spatially rescales if required.
\circlenum{$\cdot$} is the dot product.
\circlenum{$\sigma$} is the sigmoid function.
We show a simplified model with fewer layers.
}
\label{fig:model}
\end{figure}

\section{Method}
\reffig{model} illustrates our end-to-end 3D instance segmentation model \name{}.
As in Mask2Former\,\cite{Cheng22CVPR}, 
our model includes
a feature backbone~(\colorsquare{m_green}), %
a Transformer decoder~(\colorsquare{lightgray}) built from
mask modules~(\colorsquare{m_red}) and Transformer decoder layers used for query refinement~(\colorsquare{m_blue}).
At the core of the model are \emph{instance queries}, which each should represent one object instance in the scene and predict the corresponding point-level instance mask.
To that end, the instance queries are iteratively refined by the Transformer decoder (\reffig{model}, \colorsquare{lightgray})
which allows the instance queries to cross-attend to point features extracted from the feature backbone and self-attend the other instance queries.
This process is repeated for multiple iterations and feature scales,
yielding the final set of refined instance queries. 
A mask module consumes the refined instance queries together with the point features, and returns (for each query) a semantic class and
a binary instance mask based on the dot product between point features and instance queries.
Next, we describe each of these components in more detail.

\parag{Sparse Feature Backbone.}
(\reffig{model}, \colorsquare{m_green})
We use a sparse convolutional U-net backbone with a symmetrical encoder and decoder, based on the MinkowskiEngine\cite{Choy19CVPR}.
Given a colored input point cloud $P \in \mathbb{R}^{N\times 6}$ of size $N$, it is first quantized into $M_0$ voxels $V \in \mathbb{R}^{M_0\times 3}$, where each voxel is assigned the average RGB color of the points within that voxel as its initial feature. 
Next to the full-resolution output feature map $\mathbf{F}_0 \in \mathbb{R}^{M_0 \times D}$, we also extract a multi-resolution hierarchy of features from the  backbone decoder before upsampling to the next finer feature map.
At each of these resolutions $r \geq 0$ we can extract features for a set of $M_r$ voxels, which we linearly project to a fixed and common dimension $D$, yielding feature matrices $\mathbf{F}_r \in \mathbb{R}^{M_r \times D}$.
We let the queries attend to features from coarser feature maps of the backbone decoder, \ie $r \geq 1$, and use the full-resolution feature map ($r=0$) to compute the auxiliary and final per-voxel instance masks.

\parag{Mask Module.}
(\reffig{model}, \colorsquare{m_red})
Given the set of $K$ instance queries $\mathbf{X} \in \mathbb{R}^{K\times D}$, we predict a binary mask for each instance and classify each of them as one of $C$ classes or as being inactive.
To create the binary mask, we map the instance queries through an MLP $f_{\text{mask}}(\cdot)$, to the same feature space as the backbone output features.
We then compute the dot product between these \textit{instance features} and the backbone features $\mathbf{F}_0$.
The resulting similarity scores are fed through a sigmoid and thresholded at 0.5, yielding the final binary mask $\mathbf{B} \in \{0,1\}^{M \times K}$:
\begin{equation}%
    \mathbf{B} = \{b_{i,j} = [\sigma(\mathbf{F}_0  f_{\text{mask}}(\mathbf{X})^T)_{i,j} > 0.5] \}.
\end{equation}
We apply the mask module to the refined queries $\mathbf{X}$ at each Transformer layer using the full-resolution feature map $\mathbf{F}_0$, to create auxiliary binary masks for the masked cross-attention of the following refinement step.
When this mask is used as input for the masked cross-attention, we reduce the resolution according to the voxel feature resolution by average pooling.
Next to the binary mask, we predict a single semantic class per instance.
This step is done via a linear projection layer into $C+1$ dimensions, followed by a softmax.
While prior work \cite{Engelmann20CVPR,Vu22CVPR,Chen21ICCV} typically needs to obtain the semantic label of an instance via majority voting or grouping over per-point predicted semantics, this information is directly contained in the refined instance queries.

\parag{Query Refinement.}
(\reffig{model}, \colorsquare{m_blue})
The Transformer decoder starts with $K$ instance queries,
and refines them through a stack of $L$ Transformer decoder layers to a final set of accurate, scene specific instance queries by cross-attending to scene features, {and reasoning at the instance-level through self-attention.}
We discuss different types of instance queries in \refsec{impl_details}.
Each layer attends to one of the feature maps from the feature backbone using standard cross-attention:
\begin{equation}
    \mathbf{X} = \text{softmax}(\mathbf{Q}\mathbf{K}^T/\sqrt{D})\mathbf{V}.
    \label{eq:attention}
\end{equation}
To do so, the voxel features $\mathbf{F}_r \in \mathbb{R}^{M_r \times D}$ are first linearly projected to a set of keys and values of fixed dimensionality $\mathbf{K}, \mathbf{V} \in \mathbb{R}^{M_r \times D}$ and our $K$ instance queries $\mathbf{X}$ are linearly projected to the queries $\mathbf{Q} \in \mathbb{R}^{K \times D}$.
This cross-attention thus allows the queries to extract information from the voxel features.
The cross-attention is followed by a self-attention step between the queries, where the keys, values, and queries are all computed based on linear projections of the instance queries.
Without such inter-query communications, the model could not avoid multiple instance queries latching onto the same object, resulting in duplicate instance masks.
Similar to most Transformer-based approaches, we use positional encodings for our keys and queries.
We use Fourier positional encodings\cite{Tancik2020NeurIPS}
based on voxel positions.
We add the resulting positional encodings to their respective keys before computing the cross-attention. %
All instance queries are also assigned a fixed (and potentially learned) positional embedding, that is not updated throughout the query refinement process.
These positional encodings are added to the respective queries in the cross-attention, as well as to both the keys and queries in the self-attention.
Instead of using the vanilla cross-attention (where each query attends to all voxel features in one resolution)
we use a masked variant where each instance query only attends to the voxels within its corresponding intermediate instance mask $\mathbf{B}$ predicted by the previous layer.
This is realized by adding $-\infty$ to the attention matrix to all voxels for which the mask is 0.
Eq.~\ref{eq:attention} then becomes:
\begin{equation}
    \mathbf{X} = \text{softmax}(\mathbf{Q}\mathbf{K}^T/\sqrt{D} + \mathbf{B}')\mathbf{V} \text{\ with \ }     \mathbf{B}_{ij}' = -\infty \cdot [\mathbf{B}_{ij} = 0]
\end{equation}
where $[\,\cdot\,]$ are Iverson brackets.
In \cite{Cheng22CVPR}, masking out the context from the cross-attention improved segmentation.
A likely reason is that the Transformer does not need to \emph{learn} to focus on a specific instance instead of irrelevant context, but is \emph{forced} to do so by design.

In practice, we attend to the $4$ coarsest levels of the feature backbone, from coarse to fine, and do this a total of $3$ times, resulting in $L=12$ query refinement steps.
The Transformer decoder layers share weights for all $3$ iterations.
Early experiments showed that this approach preserves the performance while keeping memory requirements in bound.

\parag{Sampled Cross-Attention.}
Point clouds in a training batch typically have different point counts.
While MinkowskiEngine can handle this, current Transformer implementations rely on a fixed number of points in each batch entry. %
In order to leverage well-tested Transformer implementations, 
in this work we propose to pad the voxel features 
and mask out the attention where needed.
In case the number of voxels exceeds a certain threshold, we resort to \emph{sampling} voxel features. %
To allow instances to have access to all voxel features during cross-attention,
we resample the voxels in each Transformer decoder layer though, and use all voxels during inference.
This can be seen as a form of dropout\cite{Srivastava14JMLR}.
In practice, this procedure saves significant amounts of memory and is crucial for obtaining competitive performance.
In particular, since the proposed sampled cross-attention requires less memory, it enables training on higher-resolution voxel grids
which is necessary for achieving competitive results on common benchmarks (\eg{}, $2$\,cm voxel side-length on ScanNet\cite{Dai17CVPR}).

\subsection{Training and Implementation Details}
\label{sec:impl_details}

\parag{Correspondences.}
Given that there is no ordering to the set of instances in a scene and the set of predicted instances,
we need to establish correspondences between the two sets during training.
To that end, we use bipartite graph matching.
While such a supervision approach is not new (\eg \cite{Stewart16CVPR, Yang19NIPS}),
recently it has become more common in Transformer-based approaches \cite{Carion20ECCV,Cheng21NeurIPS,Cheng22CVPR}.
We construct a cost matrix $\mathcal{C} \in \mathbb{R}^{K \times \hat{K}}$, where $\hat{K}$ is the number of ground truth instances in a scene.
The matching cost for a predicted instance $k$ and a target instance $\hat{k}$ is given by:
\begin{equation}
\small
\mathcal{C}(k,\hat{k}) = \lambda_{\text{dice}}\mathcal{L}_\textrm{dice}(k,\hat{k}) + \lambda_{\text{BCE}}\mathcal{L}_{\textrm{BCE}_\textrm{mask}}(k,\hat{k}) + \lambda_{\text{cl}}\mathcal{L}_{\textrm{CE}_\textrm{cl}}(k,\hat{k})
\end{equation}
We set the weights to $\lambda_{\text{dice}}=\lambda_{\text{cl}}=2.0$ and $\lambda_{\text{BCE}}=5.0$.
The optimal solution for this cost assignment problem is efficiently found using the Hungarian method \cite{Kuhn55hungarian}.
After establishing the correspondences, we can directly optimize each predicted mask as follows:
\begin{equation}
\mathcal{L}_{\textrm{mask}} = \lambda_{\text{BCE}}\mathcal{L}_{\textrm{BCE}} + \lambda_{\text{dice}}\mathcal{L}_{\textrm{dice}},
\label{eq:loss_mask}
\end{equation}
where $\mathcal{L}_{\textrm{BCE}}$ is the binary cross-entropy loss (over the foreground and background of that mask) and $\mathcal{L}_{\textrm{dice}}$ is the Dice loss \cite{Deng18ECCV}.
We use the default multi-class cross-entropy loss $\mathcal{L}_{\textrm{CE}_\textrm{cl}}$ to supervise the classification.
If a mask is left unassigned, we seek to maximize the associated \emph{no-object} class, for which the $\mathcal{L}_{\textrm{CE}_\textrm{cl}}$ loss is weighted by an additional $\lambda_\text{no-obj.}=0.1$.
The overall loss for all auxiliary instance predictions after each of the $L$ layers is defined as:
\begin{equation}
\mathcal{L} = \Sigma_l^L ~ \mathcal{L}^l_{\textrm{mask}}  + \lambda_\text{cl} \mathcal{L}^l_{\textrm{CE}_\textrm{cl}}
\end{equation}

\parag{Prediction Confidence Score.}
We seek to assign a confidence to each predicted instance.
While other existing methods require a dedicated ScoreNet\cite{Jiang20CVPR} which is trained to estimate the intersection over union with the ground truth instances, we directly obtain the confidence scores from the refined query features and point features as in Mask2Former\cite{Cheng22CVPR}.
We first select the queries with a dominant semantic class, for which we obtain the class confidence based on the softmax output $c_\textrm{cl} \in [0,1]$, which we additionally multiply with a mask based confidence:
\begin{equation}
c = c_\textrm{cl} \cdot (\Sigma_i^M m_i \cdot [m_i > 0.5]) / ( \Sigma_i^M [m_i > 0.5]),
\end{equation}
where $m_i \in [0,1]$ is the instance mask confidence for the $i^{th}$ voxel given a single query.
In essence, this is the mean mask confidence of all voxels falling inside of the binarized mask \cite{Cheng22CVPR}.
For an instance prediction to have a high confidence, it needs both a confident classification among $C$ classes, and a mask that predominantly consists of high-confidence voxels.

\parag{Query Types.}
Methods like DETR\cite{Carion20ECCV} or Mask2Former\cite{Cheng21NeurIPS, Cheng22CVPR} use parametric queries. %
During training both the instance query features and the corresponding positional encodings are learned.
This thus means that during training the set of $K$ instance queries has to be optimized in such a way that it can cover all instances present in a scene during inference.

Misra~\etal\cite{Misra21ICCV} propose to initialize queries with sampled point coordinates from the input point cloud based on farthest point sampling.
Since this initialization does not involve learned parameters, they are called \emph{non-parametric} queries.
Interestingly, the instance query features are initialized with zeros and only the 3D position of the sampled points is used to set the corresponding positional encoding.
We also experiment with a variant where we use sampled point features as instance query features.
Similar to \cite{Misra21ICCV}, we observe improved performance when using non-parametric queries although less pronounced.
The key advantage of non-parametric queries is that, during inference, we can sample a different number of queries than during training.
This provides a trade-off between inference speed and performance, without the need to retrain the model when using more instance queries.

\parag{Training Details.}
The feature backbone is a Minkowski Res16UNet34C\cite{Choy19CVPR}.
We train for $600$\,epochs using AdamW\cite{Loshchilov19ICLR} and a one-cycle learning rate schedule\cite{Smith19AIMLMDOA} with a maximal learning rate of $10^{-4}$. 
Longer training times (1000 epochs) did not further improve results.
One training on 2\,cm voxelization takes $\sim$78 hours on an NVIDIA A40 GPU.
We perform standard data augmentation: horizontal flipping, random rotations around the z-axis, elastic distortion\cite{Ronneberger2015MICCAI} and random scaling.
Color augmentations include jittering, brightness and contrast augmentations.
During training on ScanNet, we reduce memory consumption by computing the dot product between instance queries and aggregated point features within segments (obtained from a graph-based segmentation\cite{Felzenszwalb04IJCV}, similar to OccuSeg\cite{Han20CVPR} or Mix3D\cite{Nekrasov2021ThreeDV}).
Wrongly merged instances can be separated using connected components\cite{Ester96KDD} (\refsec{eval_quali}).

\begin{table}[t]
\vspace{5pt}
	\caption{
	    \textbf{3D Instance Segmentation Scores on ScanNet\,v2\,\cite{Dai17CVPR}.}
	    We report mean average precision (mAP) with different IoU threshold over 18 classes on the ScanNet validation and test set.
	    The inference speed is averaged over the validation set and computed on a TITAN X GPU~(\cf~\cite{Vu22CVPR}), excluding postprocessing. Test scores accessed on 13. September 2022.}
    \small
    \setlength{\tabcolsep}{2px}
    \begin{tabularx}{\linewidth}{lYYcYYcc}
    \toprule
    & \multicolumn{2}{c}{\bf ScanNet Val} && \multicolumn{2}{c}{\bf ScanNet Test} && \multirow{3}{*}{ \parbox{0.9cm}{\centering {\bf Runtime} (in ms) }}\\
    \cmidrule{2-3} \cmidrule{5-6}
    Method                          &     mAP  & mAP$_{50}$ && mAP & mAP$_{50}$ &&   \\ \midrule
    SGPN \cite{Wang18CVPR}         &      --   & --       && 4.9   & 14.3         &&158439\\
    GSPN \cite{yi2019gspn}         &       19.3 & 37.8  && --    & 30.6               &&12702\\
    3D-SIS \cite{Hou19CVPR}        &          --   & 18.7   && 16.1    & 38.2           &&--\\
    MASC \cite{Liu19ARXIV}         &          --   & --     && 25.4  & 44.7           &&--\\
    3D-Bonet \cite{Yang19NIPS}     &        --   & --    && 25.3  & 48.8         &&9202\\
    MTML \cite{Lahoud19ICCV}       &          20.3 & 40.2   && 28.2  & 54.9           &&--\\
    3D-MPA \cite{Engelmann20CVPR}  &          35.5 & 59.1  && 35.5 & 61.1             &&--\\
    DyCo3D \cite{He21CVPR}         &          35.4 & 57.6 && 39.5  & 64.1            &&--\\
    PointGroup \cite{Jiang20CVPR}  &         34.8 & 56.7  && 40.7 & 63.6            &&452\\
    MaskGroup \cite{Zhong22Arxiv}  &          42.0 & 63.3   && 43.4 & 66.4            &&--\\
    OccuSeg \cite{Han20CVPR}       &        44.2 & 60.7  && 48.6 & 67.2           &&1904\\
    SSTNet \cite{Liang21ICCV}      &         49.4 & 64.3 && 50.6  & 69.8           &&428\\
    HAIS \cite{Chen21ICCV}         &   43.5 & 64.1 && 45.7  & 69.9     &&\textbf{339}\\
    SoftGroup \cite{Vu22CVPR}     &         46.0 & 67.6  && 50.4 & 76.1 && 345\\
    \name{} (Ours)                 &    \textbf{55.2} & \textbf{73.7} && \textbf{56.6} & \textbf{78.0} && \textbf{339} \\
    \bottomrule
    \end{tabularx}
	\label{tab:scores_scannet}
\end{table}

\section{Experiments}
In this section,
we compare \name{} with prior state-of-the-art on four publicly available 3D indoor and outdoor datasets (\refsec{eval_sota}).
Then, we provide analysis experiments on the proposed model investigating query types and the impact of the number of query refinement steps as well as the number of queries during inference. (\refsec{eval_analysis}).
Finally, we show qualitative results and discuss limitations (\refsec{eval_quali}).

\subsection{Comparing with State-of-the-Art Methods}
\label{sec:eval_sota}

\parag{Datasets and Metrics.}
We evaluate \name{} on four publicly available 3D instance segmentation datasets.

\textit{ScanNet}\cite{Dai17CVPR} is a richly-annotated dataset of 3D reconstructed indoor scenes.
It contains hundreds of different rooms showing a large variety of room types such as hotels, libraries and offices.
The provided splits contain 1202 training, 312 validation and 100 hidden test scenes.
Each scene is annotated with semantic and instance segmentation labels covering 18 object categories.
The benchmark evaluation metric is mean average precision (mAP).
\textit{ScanNet200}\cite{rozenberszki22ECCV} extends the original ScanNet scenes with an order of magnitude more classes.
ScanNet200 allows to test an algorithm's performance under the natural imbalance of classes,
particularly for challenging long-tail classes such as \textit{coffee-kettle} and \textit{potted-plant}.
We keep the same train, validation and test splits as in the original ScanNet dataset.
\textit{S3DIS}\cite{Armeni16CVPR} is a large-scale indoor dataset showing six different areas from three different campus buildings.
It contains 272 scans and is also annotated with semantic instance masks over 13 different classes.
We follow the common splits and evaluate on Area-5 and 6-fold cross validation.
We report scores using the mAP metric from ScanNet and mean precision/recall at IoU threshold 50\% (mPrec$_{50}$/mRec$_{50}$) as initially introduced by ASIS\cite{Wang19CVPR}.
Unlike mAP, this metric does not consider confidence scores,
therefore we filter out instance masks with a prediction confidence score below 80\% to avoid excessive false positives.
\textit{STPLS3D}\cite{Chen22Arxiv} is a synthetic outdoor dataset closely mimicking the data generation process of aerial photogrammetry point clouds. %
25 urban scenes totalling 6 \,km$^2$ are densely annotated with 14 instance classes.
We follow the common splits \cite{Chen22Arxiv, Vu22CVPR}.

\parag{Results} are summarized in \reftab{scores_scannet} (ScanNet), \reftab{scores_stanford} (S3DIS), \reftab{scores_scannet200_stpls3d} (\emph{left}, ScanNet200) and \reftab{scores_scannet200_stpls3d} (\emph{right}, STPLS3D).
\name{} outperforms prior work by a large margin
on the most challenging metric mAP by at least \textbf{6.2\,mAP} on ScanNet, \textbf{6.2\,mAP} on S3DIS, \textbf{10.8\,mAP} on ScanNet200 and {\textbf{11.2\,mAP}} on STPLS3D.
As in \cite{Chen21ICCV, Vu22CVPR}, we also report scores for models pre-trained on ScanNet and fine-tuned on S3DIS. 
For \name{}, pre-training improves performance by {1.2\,mAP} on Area\,5.
Mask3D's strong performance on indoor and outdoor datasets as well as its ability to work under challenging class imbalance settings \emph{without} inherent modifications to the architecture or the training regime highlights its generality.
Trained models are available at: {\small \url{https://github.com/JonasSchult/Mask3D}}

\begin{table}[t]
\vspace{5pt}
\caption{
    \textbf{3D Instance Segmentation Scores on S3DIS\,\cite{Armeni16CVPR}.}
    We report mean average precision (mAP) with different IoU threshold (as in \cite{Dai17CVPR}) as well as mean precision (mPrec) and mean recall (mRec) with 50\% IoU threshold (as in \cite{Wang19CVPR}) over 13 classes on S3DIS Area 5 and 6-fold cross validation.
    Scores in light gray are pre-trained on ScanNet\,\cite{Dai17CVPR} and fine-tuned on S3DIS\,\cite{Armeni16CVPR}.
}
	\label{tab:scores_stanford}
\small
\setlength{\tabcolsep}{1.2px}
\begin{tabu}{lccccccccc} 
\toprule
& \multicolumn{4}{c}{\bf S3DIS Area 5} & & \multicolumn{4}{c}{\bf S3DIS 6-fold CV} \\
\cmidrule{2-5} \cmidrule{7-10}
Method                          & AP & AP$_{50}$ & Prec$_{50}$ & Rec$_{50}$ && AP & AP$_{50}$ & Prec$_{50}$ & Rec$_{50}$ \\ \midrule
SGPN \cite{Wang18CVPR}             & --   & --   & 36.0 & 28.7 && --   & --   & 38.2 & 31.2 \\
ASIS \cite{Wang19CVPR}                       & --   & --   & 55.3 & 42.4 && --   & --   & 63.6 & 47.5 \\
3D-Bonet \cite{Yang19NIPS}         & --   & --   & 57.5 & 40.2 && --   & --   & 65.6 & 47.6 \\
OccuSeg \cite{Han20CVPR}           & --   & --   & --   & --   && --   & --   & 72.8 & 60.3 \\
3D-MPA \cite{Engelmann20CVPR}      & --   & --   & 63.1 & 58.0 && --   & --   & 66.7 & 64.1 \\
PointGroup \cite{Jiang20CVPR}      & --   & 57.8 & 61.9 & 62.1 && --   & 64.0 & 69.6 & 69.2 \\
DyCo3D \cite{He21CVPR}             & --   & --   & 64.3 & 64.2 && --   & --   & --   & --   \\
MaskGroup \cite{Zhong22Arxiv}      & --   & 65.0 & 62.9 & 64.7 && --   & 69.9 & 66.6 & 69.6 \\
SSTNet \cite{Liang21ICCV}          & 42.7 & 59.3 & 65.5 & 64.2 && 54.1 & 67.8 & \textbf{73.5} & 73.4 \\
\name{} (Ours)                      & \textbf{56.6} & \textbf{68.4} & \textbf{68.7} & \textbf{66.3}   && \textbf{64.5}   & \textbf{75.5}   & 72.8   & \textbf{74.5}   \\
\midrule
\rowfont{\color{gray}}
HAIS \cite{Chen21ICCV}             & --   & --   & 71.1 & 65.0 && --   & --   & 73.2 & 69.4 \\
\rowfont{\color{gray}}
SoftGroup \cite{Vu22CVPR}    & 51.6 & 66.1 & 73.6 & \textbf{66.6} && 54.4 & 68.9 & 75.3 & \textbf{69.8} \\
\rowfont{\color{gray}}
\name{} (Ours)          & \textbf{57.8} & \textbf{71.9}   & \textbf{74.3}   & 63.7   && \textbf{61.8}   & \textbf{74.3} & \textbf{76.5} & 66.2   \\
\bottomrule
\end{tabu}
\end{table}

\begin{table}[t]
\centering
	\caption{
	    \textbf{3D Instance Segmentation Scores on ScanNet200\,\cite{rozenberszki22ECCV} and STPLS3D\,\cite{Chen22Arxiv}.} We report mean average precision (mAP) with different IoU threshold over 14 classes on the STPLS3D test set.
	    Hidden test scores accessed on 13. September 2022.
	}
\begin{minipage}{.48\linewidth}
    \centering
        \small
        \setlength{\tabcolsep}{0.5px}
        \begin{tabularx}{\linewidth}{lYYY}
    \toprule
    & \multicolumn{3}{c}{\textbf{ScanNet 200}}\\
    \cmidrule{2-4}
    Method                           & head & com & tail\\ \midrule
    CSC\cite{Hou21CVPR}  & 22.3 & 8.2 & 4.6\\
    Mink34D\cite{Choy19CVPR} & 24.6 & 8.3 & 4.3 \\
    LGround\cite{rozenberszki22ECCV} & 27.5 & 10.8 & 6.0\\
    \name{} (Ours) & \textbf{38.3} & \textbf{26.3} & \textbf{16.8} \\
    \bottomrule
    \end{tabularx}
	\label{tab:scores_scannet200}
\label{tab:query_types}
\end{minipage}%
\hfill
\begin{minipage}{.48\linewidth}
    \centering
    \small
            \setlength{\tabcolsep}{0.0px}
    \begin{tabularx}{\linewidth}{lYY}
    \toprule
    & \multicolumn{2}{c}{\bf STPLS3D}\\
    \cmidrule{2-3}
    Method                          &     mAP  & mAP$_{50}$ \\ \midrule
    PointGroup \cite{Jiang20CVPR}  &         23.3 & 38.5\\
    HAIS \cite{Chen21ICCV}         &   35.1 & 46.7\\
    SoftGroup \cite{Vu22CVPR}     &         46.2 & 61.8\\
    \name{} (Ours)                 & \textbf{57.3}    & \textbf{74.3} \\
    \bottomrule
    \end{tabularx}
	\label{tab:scores_stpls3d}
\end{minipage}
\label{tab:scores_scannet200_stpls3d}
\end{table}

\subsection{Analysis Experiments}
\label{sec:eval_analysis}
\parag{Query Types.}
\begin{table}[t]
\caption{\textbf{Ablations.}
\textbf{a)}
We explore two variants for query positions and features.
Parametric queries \circlenum{1} are learned during training.
Non-parametric queries consist of FPS point positions \circlenum{2} and potentially their features \circlenum{3}, resembling scene-specific queries.
\textbf{b)} We optimize the instance mask prediction using the binary cross-entropy loss $\mathcal{L}_{CE}$ and the dice loss $\mathcal{L}_{dice}$.
A weighted combination of dice and cross-entropy loss results in best performance.}
\centering
\begin{minipage}{.49\linewidth}
    \centering
    \setlength{\tabcolsep}{1.0px}
\small
\begin{tabular}{cllccc} 
	    \toprule
		& Position & Features && mAP \\
		\midrule
		\circlenum{1} & Param. & Param. && $39.7${\scriptsize $\pm$$0.7$} %
		\\
		\circlenum{2} & FPS & Zeros && $\mathbf{40.6}${\scriptsize$\pm$$0.3$} & %
        \\
		\circlenum{3} & FPS & Point Feat. && $38.4${\scriptsize $\pm$$0.3$}
		\\
		\bottomrule
	\end{tabular}%

 \vfill
    \vspace{3px}
    \textbf{a)} Query Type
\label{tab:query_types}
\end{minipage}%
\begin{minipage}{.49\linewidth}
    \centering
    \setlength{\tabcolsep}{1.0pt}
\small
\begin{tabular}{cccc} 
    \toprule
	$\mathcal{L}_{\mathrm{dice}}$ & $\mathcal{L}_{\mathrm{BCE}}$ && mAP
   \\
	\midrule
	\xmark & \cmark && $27.0${\scriptsize $\pm$$0.6$}
   \\
	\cmark & \xmark && $38.0${\scriptsize $\pm0.3$}
 \\
	\cmark & \cmark && $\mathbf{40.6}${\scriptsize $\pm0.3$}
 \\

	\bottomrule
\end{tabular}%
 \vfill
    \vspace{3px}
    \textbf{b)} Mask Loss
\label{tab:mask_loss}
\end{minipage}
\label{tab:ablations}
\end{table}
\name{} iteratively refines instance queries by attending to %
voxel features (\reffig{model}, \colorsquare{lightgray}).
We distinguish two types of query initialization prior to attending to voxel features: \circlenum{1} parametric and \circlenum{2}-\circlenum{3} non-parametric initial queries.
Parametric refers to \emph{learned} positions and features \cite{Carion20ECCV}, %
while non-parametric refers to point positions sampled with \emph{furthest point sampling} (FPS)\cite{Qi17NeurIPS}.
When selecting query positions with FPS, we can either initialize the queries to zero (\circlenum{2}, as in 3DETR\cite{Misra21ICCV}) or use the point features at the sampled position \circlenum{3}.
\reftab{ablations} (\emph{left}) shows the effects of using parametric or non-parametric queries on ScanNet validation ($5$\,cm).
In line with \cite{Misra21ICCV}, we see that non-parametric queries\,\circlenum{2} outperform parametric queries\,\circlenum{1}.
Interestingly, \circlenum{3} results in degraded performance compared to both parametric\,\circlenum{1} and position-only non-parametric queries\,\circlenum{2}.

\parag{Number of Queries and Decoders.}
\begin{figure}[t]
    \centering
    \includegraphics[width=\linewidth]{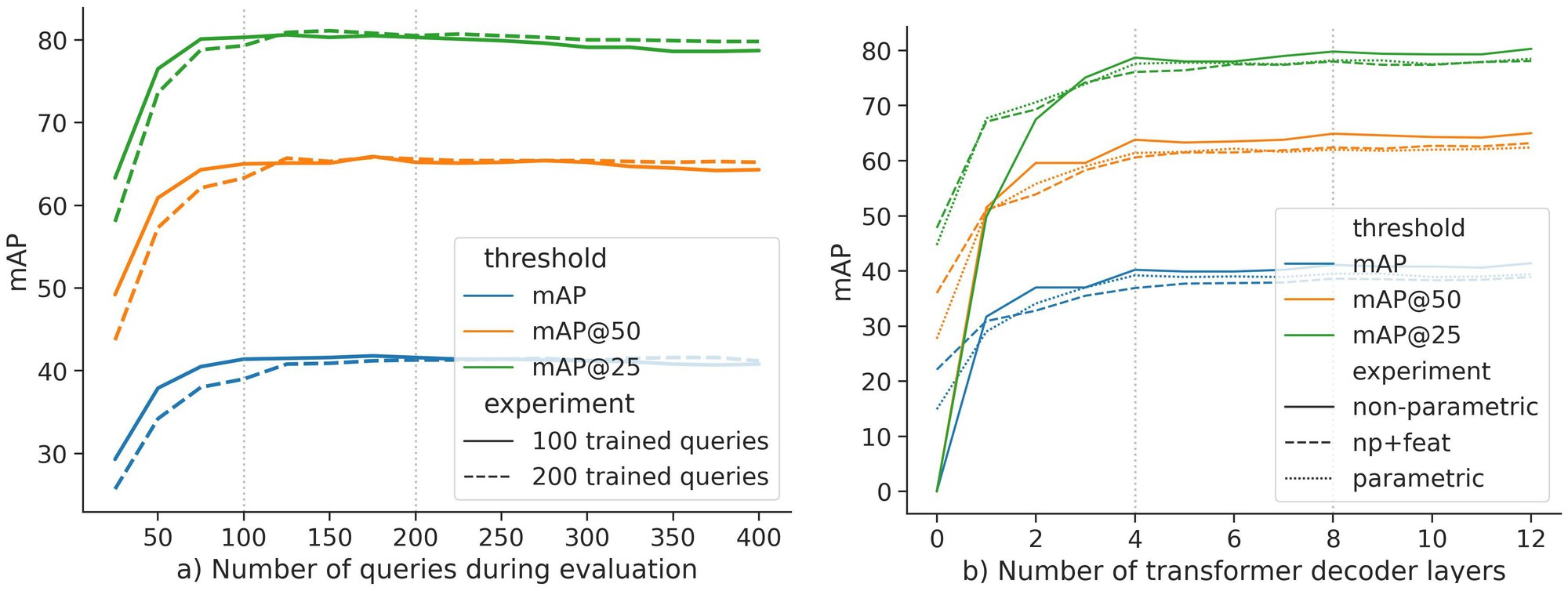}
\caption{Number of queries and decoder layers.
}
\label{fig:eval_num_queries_decoders}
\end{figure}
We analyze the effect of varying numbers of queries $K$ during inference on models trained with $K$\,$=$\,$100$ and $K$\,$=$\,$200$ non-parametric queries sampled with FPS.
By increasing $K$ from 100 to 200 during training, we observe a slight increase in performance (\reffig{eval_num_queries_decoders}, \emph{left}) at the cost of additional memory.
When evaluating with fewer queries than trained with,
we observe reduced performance but faster runtime.
When evaluating with more queries than trained with,
we observe slightly improved performance, typically less than $1$\,mAP.
Our final model uses $K$\,$=$\,$100$ due to memory constraint when using 2\,cm voxels in the feature backbone.
In this study, we report scores using 5\,cm on ScanNet validation. 
We also analyse the mask quality that we obtain after each Transformer decoder layer in our trained model (\reffig{eval_num_queries_decoders}, \emph{right}).
We see a rapid increase up to 4 layers, then the quality increases a bit slower. 

\parag{Mask Loss.}
The mask module (\reffig{model},\,\colorsquare{m_red}) generates instance heatmaps for every instance query.
After Hungarian matching, the corresponding ground truth mask is used to compute the mask loss $\mathcal{L}_{\mathrm{mask}}$.
The binary cross entropy loss $\mathcal{L}_{\mathrm{BCE}}$ is the obvious choice for binary segmentation tasks.
However, it does not perform well under large class imbalance (few foreground mask points, many background points).
The Dice loss $\mathcal{L}_{\mathrm{dice}}$ is specifically designed to address such data imbalance.
\reftab{ablations} (\emph{right}) shows scores on ScanNet validation for combinations of both losses.
While $\mathcal{L}_{\mathrm{dice}}$ improves over $\mathcal{L}_{\mathrm{BCE}}$,
we observe an additional improvement by training our model with a weighted sum of both losses (\refeq{loss_mask}).%

\subsection{Qualitative Results and Limitations}
\label{sec:eval_quali}

\begin{figure}[t]
\includegraphics[width=1.0\linewidth]{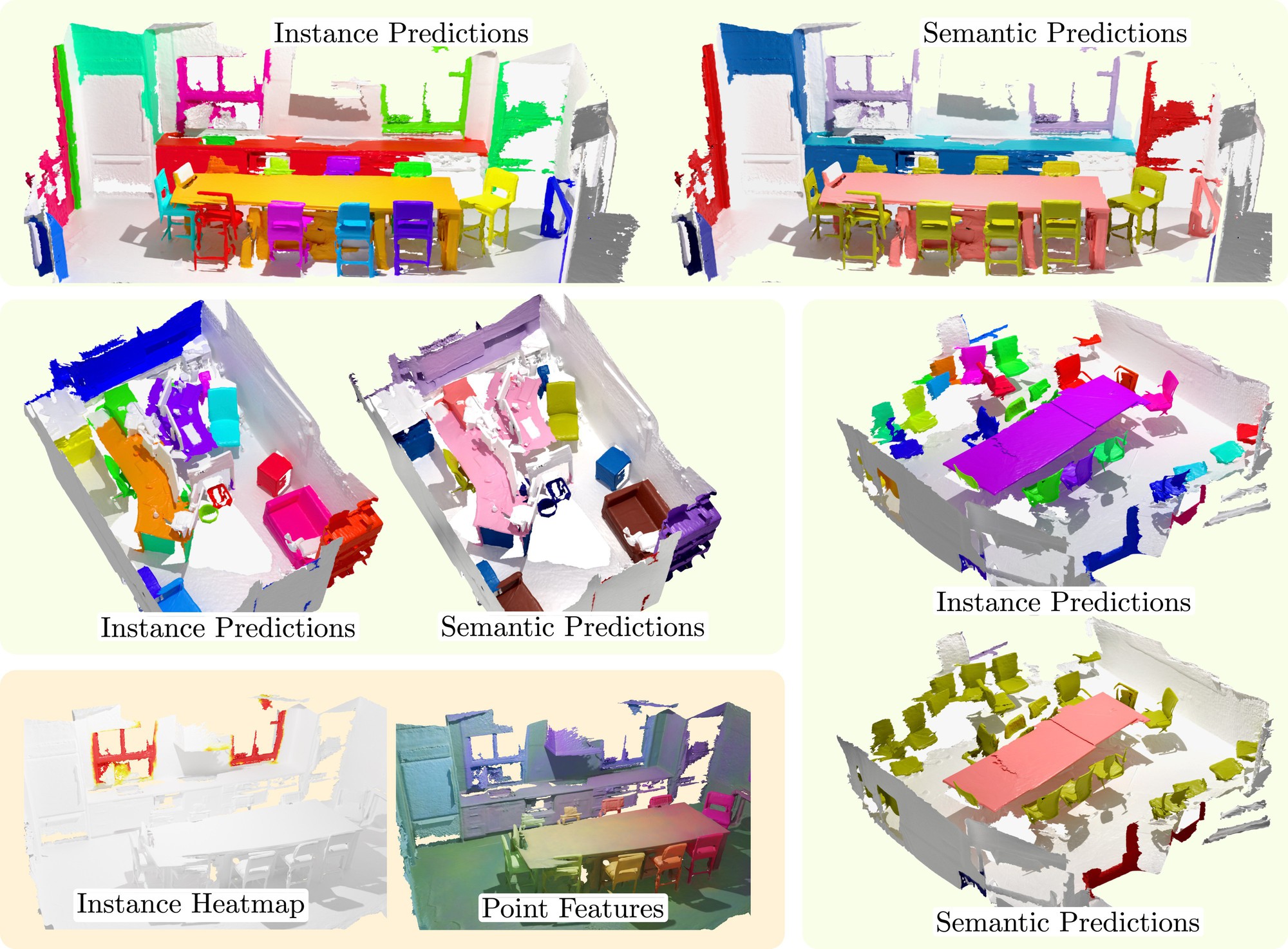}
\caption{\textbf{Qualitative Results on ScanNet.}
We show pairs of predicted instance masks and predicted semantic labels. On the bottom left, we show the heatmap of a failure case of two windows that are wrongly assigned to a single instance.
The corresponding point features are visualized as RGB after projecting them to 3D using PCA.
\vspace{-10px}
}
\label{fig:qualitatve_grid}
\end{figure}

\reffig{qualitatve_grid} shows several representative examples of \name{} instance segmentation results on ScanNet.
The scenes are quite
diverse and present a number of challenges, including clutter, scanning artifacts and numerous similar objects.
Still, our
model shows quite robust results.
There are still limitations in our model though.
A systematic mistake that we observed are merged instances that are far apart (see \reffig{qualitatve_grid}, \emph{bottom left}).
As the attention mechanism can attend to the full point cloud, it can happen that two objects with similar semantics and geometry expose similar learned point features and are therefore combined into one instance even if they are far apart in the scene.
This is less likely to happen with methods that explicitly encode geometric priors.

\section{Conclusion}
\label{sec:conclusion}
In this work, we have introduced \name{}, 
for 3D semantic instance segmentation.
\name{} is based on Transformer decoders, and learns instance queries that, combined with learned point features, directly predict semantic instance masks without the need for hand-selected voting schemes or hand-crafted grouping mechanisms.
We think that \name{} is an attractive alternative to current voting-based approaches and expect to see follow-up work along this line of research.

{\small
\parag{\small{Acknowledgments:}}
This work is supported by the ERC Consolidator Grant DeeViSe (ERC-2017-CoG-773161), SNF Grant
200021 204840, compute resources from RWTH Aachen University (rwth1238)
and the ETH AI Center post-doctoral fellowship.
We additionally thank Alexey Nekrasov, Ali Athar and István Sárándi for helpful discussions and feedback.
}

\bibliographystyle{plain}
\balance
\bibliography{abbrev,egbib}

\clearpage

\twocolumn[{%
\renewcommand\twocolumn[1][]{#1}%
\vspace{0.1cm}
\begin{center}
\textbf{\Large Mask3D: Mask Transformer for 3D Semantic Instance Segmentation\\[0.2cm]
\textit{Supplementary Material}}
\end{center}
\vspace{1.7cm}
}]

\setcounter{section}{0}

\normalsize{

\section{Implementation Details}

\parag{S3DIS Specific Details.}
As S3DIS\,\cite{Armeni16CVPR} contains a few very large spaces, \eg lecture halls, and also provides a very high point density, scenes can exceed several millions of points.
We therefore resort to training on $6$m$\times6$m blocks randomly cropped from the ground plane to keep the memory requirements in bounds.
As \name{} thus effectively sees less data in each epoch, we train for $1000$ epochs.
However, during test, we disable cropping and infer full scenes.

\parag{STPLS3D Specific Details.}
As STPLS3D's evaluation protocol\,\cite{Chen22Arxiv} evaluates on $50$m$\times50$m blocks evenly cropped from the full city scene, instances are potentially separated into multiple blocks.
We therefore feed slightly larger $54$m$\times54$m blocks in our model but only keep the relevant predicted instances of the $50$m$\times50$m block.
This approach achieves significantly better results, usually roughly $1.2$\,mAP.

\parag{Model Details.}
Figure~\ref{fig:full_model} shows our full model.
Unlike the figure in the main paper, this shows the complete model, including all backbone feature levels and all query refinement steps in the Transformer decoder.
We deploy a Minkowski Res16UNet34C\,\cite{Choy19CVPR} and obtain feature maps $\mathbf{F}_i$ from all of its 5 scales.
The feature maps have $(96, 96, 128, 256, 256)$ channels (sorted from fine to coarse).
As the Transformer decoder expects a feature dimension of $128$, we apply a non-shared linear projection after each $\mathbf{F}_i$ to map the features to the expected dimension.
Furthermore, we employ a modified Transformer decoder by Mask2Former\,\cite{Cheng22CVPR} (swapped cross- and self-attention) leveraging an 8-headed attention and a feedforward network with $1024$-dimensional features.
For each intermediate feature map $\mathbf{F}_i$ with $i>0$, we instantiate a dedicated decoder layer.
We attend to the backbone features 3 times with Transformer decoders with shared weights.
In all our experiments, we use $100$ instance queries.
Following Misra \etal\,\cite{Misra21ICCV}, the query positions are calculated from Fourier positional encodings based on relative voxel positions scaled to $[-1,1]$.
We do not use Dropout.

\begin{figure}
\centering
\includegraphics[width=0.4\textwidth]{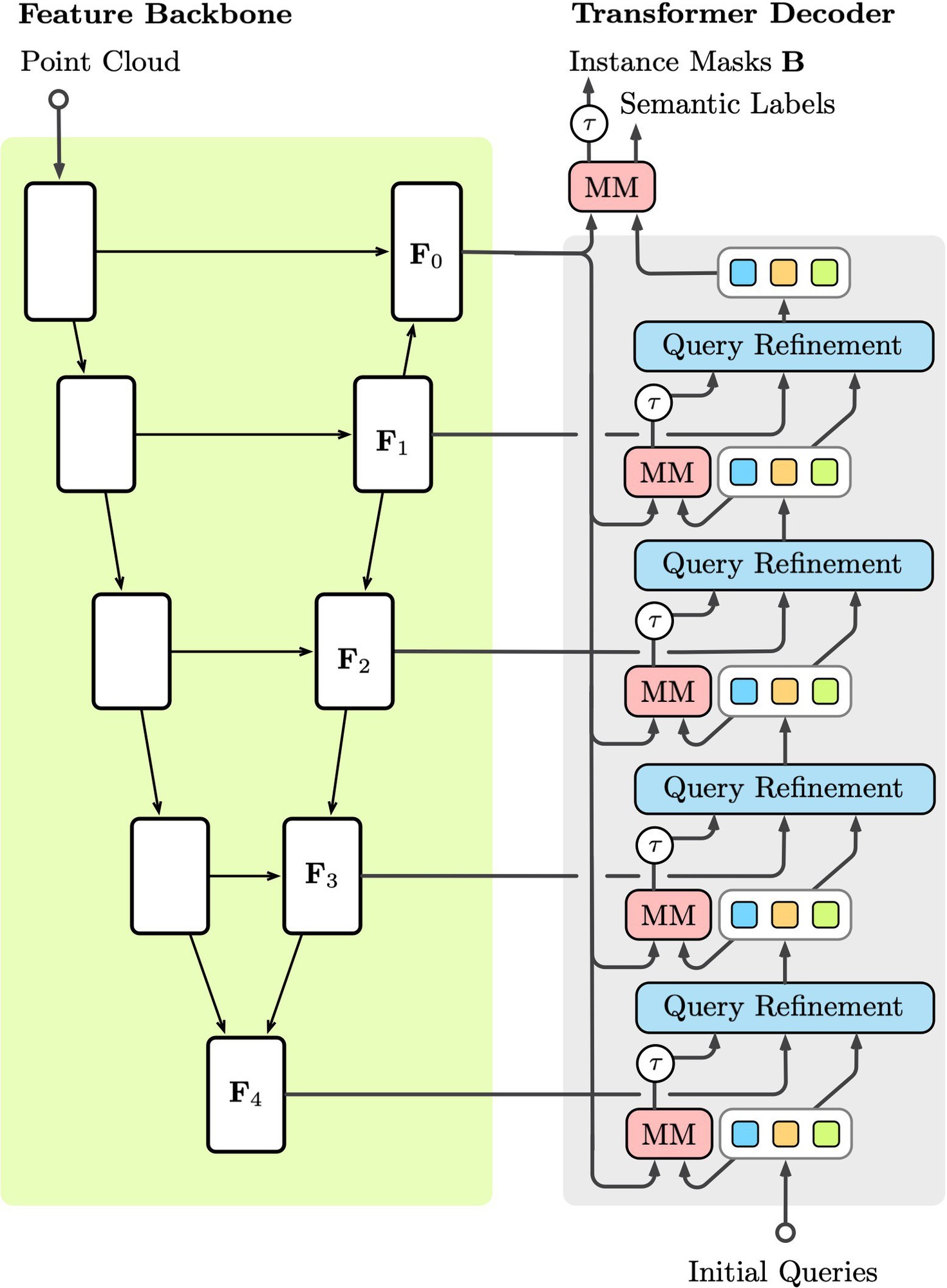}
\caption{\textbf{Illustration of the full \name{} model.}
In the main paper, we showed a simplified version of our model with fewer hierarchical feature levels in the feature backbone (shown in green) and fewer query refinement layers (blue).
The feature backbone outputs point features in 5 scales, while the Transformer decoder iteratively refines the instance queries.
Given point features and instance queries, the mask module predicts for each query a semantic class and an instance heatmap, which (after thresholding) results in a binary instance mask.
}
\label{fig:full_model}
\end{figure}

\parag{Comparison Feature Backbones.}
\begin{table}[t]
    \centering
	\caption{\textbf{Feature Backbones.}
 We experimented with convolutional and transformer-based feature backbones (\cf \reffig{full_model}, \colorsquare{m_green}).
	}
    
	\begin{tabularx}{\linewidth}{ccYY}
		\toprule
		Backbone Name & Backbone Param. & mAP & mAP$_{50}$\\
		\midrule
        StratifiedFormer\,\cite{Lai22CVPR} & 18,798,662 & 31.1 & 54.6 \\
        Res16UNet18B\,\cite{Choy19CVPR} & 17,204,660 & 40.0 & 63.7 \\
        Res16UNet34C\,\cite{Choy19CVPR} & 37,856,052 & \textbf{40.9} & \textbf{64.4} \\
		\bottomrule
	\end{tabularx}%
	\label{tab:feature_backbones}
\end{table}
As an additional candidate for non-convolution-based backbones, we deploy the recent StratifiedFormer\,\cite{Lai22CVPR} which is a Transformer-based feature backbone.
The resulting scores are reported in \reftab{feature_backbones}.
The experiment with the StratifiedFormer shows encouraging results but does not yet reach the performance of the sparse convolutional backbone.
However, the experiment clearly shows that our model also runs on different types of feature backbones.
We also report scores of another voxel-based feature backbone (Minkowski Res16UNet18B) that is significantly smaller than our original backbone (Minkowski Res16UNet34C) to show robustness towards model size on ScanNet validation.
We find that the smaller feature backbone works comparably to the bigger Res16UNet34C.
This shows that Mask3D does not overly rely on the specific voxel-based feature extractor.

\parag{Model sizes.}
\begin{table}[t]
    \centering
	\caption{\textbf{Model sizes.}
    We compare Mask3D's model size against recent top-performing methods.
    For all models, most parameters are in the feature backbone and only a small fraction is in the instance segmentation specific part of the models. 
	}
	\begin{tabularx}{\linewidth}{lc|YY}
		\toprule
		Model Name & All Params. & Backbone & Other \\
		\midrule
       HAIS\,\cite{Chen21ICCV} & 30.856M & 30.118M & 0.738M \\
       SoftGroup\,\cite{Vu22CVPR} & 30.858M & 30.118M & 0.740M \\
       Mask3D (Ours) & 39.617M & 37.856M & 1.761M \\
       Mask3D (Ours -- small) & 18.958M & 17.205M & 1.753M \\
		\bottomrule
	\end{tabularx}%
	\label{tab:model_sizes}
\end{table}
\reftab{model_sizes} shows the model size of Mask3D and two recent top-performing baselines HAIS\,\cite{Chen21ICCV} and SoftGroup\,\cite{Vu22CVPR} obtained from the official code releases.
The most parameters, by far (>$90\%$), are due to the feature-learning backbones (\reffig{full_model}, \colorsquare{m_green}).
In comparison, the remaining number of parameters (including the transformer-decoder) is very small (<$10\%$).
In absolute numbers, the proposed transformer-decoder is larger than the other parts of the baseline methods but still small compared to the size of the feature backbones.

To verify that the improved performance of Mask3D does not originate from more model parameters, we ran an additional experiment with a smaller feature backbone (Res16UNet18B).
The smaller feature backbone results in comparable segmentation performance ($40.9$ vs. $40.0$\,mAP) evaluated on ScanNet validation 5\,cm.
Additional feature backbones are analyzed in \reftab{feature_backbones}.

\subsection{Comparison to SoftGroup}
In the following, we qualitatively compare \name{} with SoftGroup\,\cite{Vu22CVPR}, the currently best performing voting-based 3D instance segmentation approach.
We highlight two error cases for SoftGroup and show our \name{} for comparison in \reffig{qualitative_comparison_softgroup}.
\begin{figure*}[t]%
    \centering%
    
    \includegraphics[width=\textwidth]{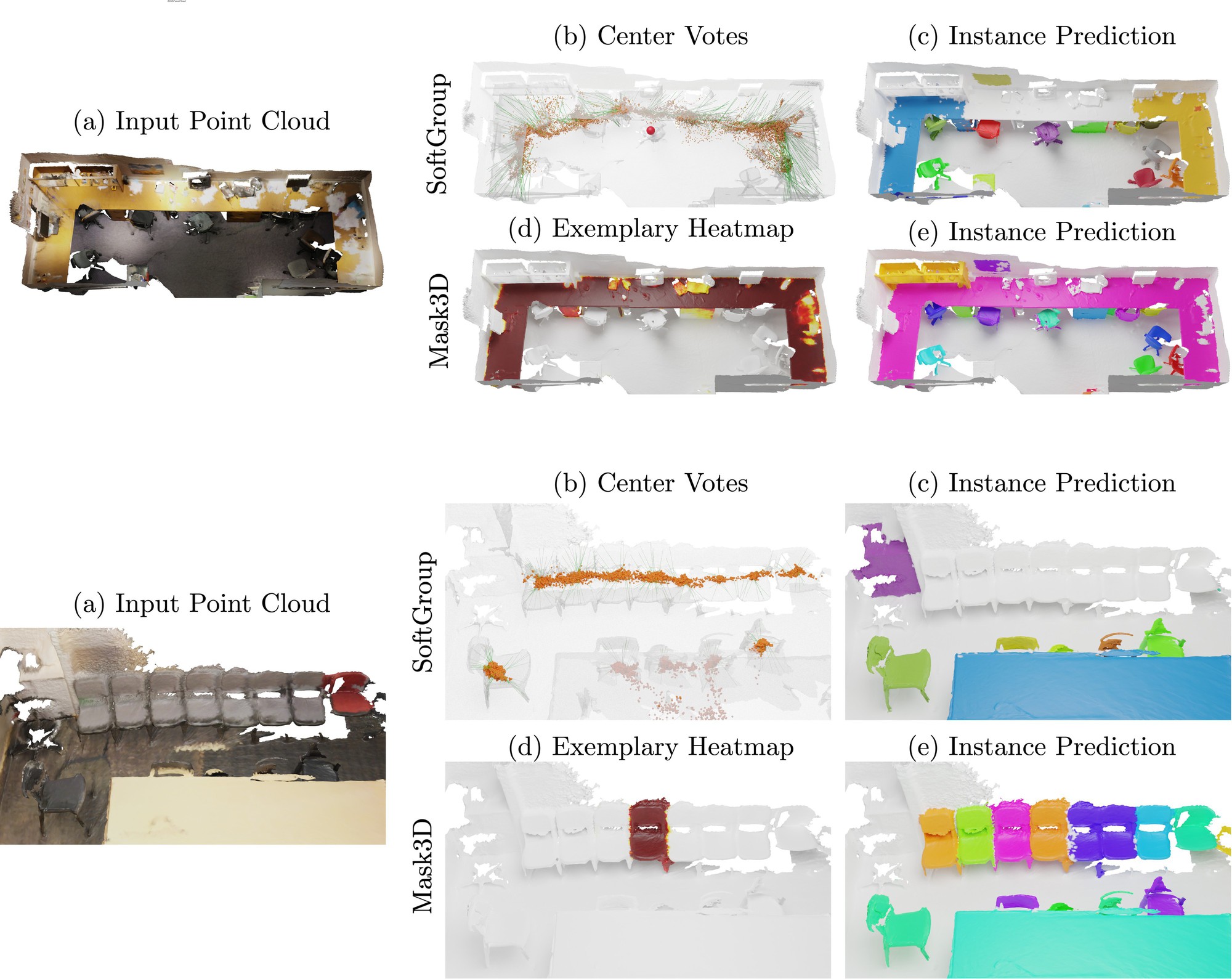}
    
    \caption{%
        \textbf{Qualitative Comparison to SoftGroup\,\cite{Vu22CVPR}.}
        We compare \name{} with the current top-performing voting-based approach SoftGroup.
        The top example shows a scene containing a single large U-shaped table, see (e) in pink. SoftGroup is based on center-voting and tries to predict the instance center, shown in (b) in red. However, predicting centers of such very large non-convex shapes can be difficult for voting-based approaches. Indeed, SoftGroup fails to correctly segment the table and returns two partial instances (c). Our \name{}, on the other side, does not rely on hand-selected geometric properties such as centers and can handle arbitrarily shaped and sized objects. It correctly predicts the tables instance mask (e).
        In the {bottom} example, we see that SoftGroup has difficulties to predict precise centers for multiple chairs located next to each other (b). As a result, the manually tuned grouping mechanism aggregates them all into one big instance which is later discarded by the refinement step. It therefore misses to segment all eight chairs (c).  \name{} does not rely on hand-crafted grouping mechanisms and can successfully segment most of the chairs.
    }
    \label{fig:qualitative_comparison_softgroup}
\end{figure*}

\parag{Density-Based Clustering.}
\begin{figure*}[t]%
    \centering%
    \newcommand{\tableline}[1]{\includegraphics[width=0.33\textwidth]{figures/#1_rgb.jpg}&\includegraphics[width=0.33\textwidth]{figures/#1_instances.jpg}&\includegraphics[width=0.33\textwidth]{figures/#1_heatmap.jpg}\\}%
    \setlength{\tabcolsep}{0pt}%
    \begin{tabularx}{\textwidth}{YY}%
    \textbf{(a)} RGB Point Cloud & \textbf{(b)} \name{} (ours) w/o DBSCAN \\
    \includegraphics[width=0.49\textwidth]{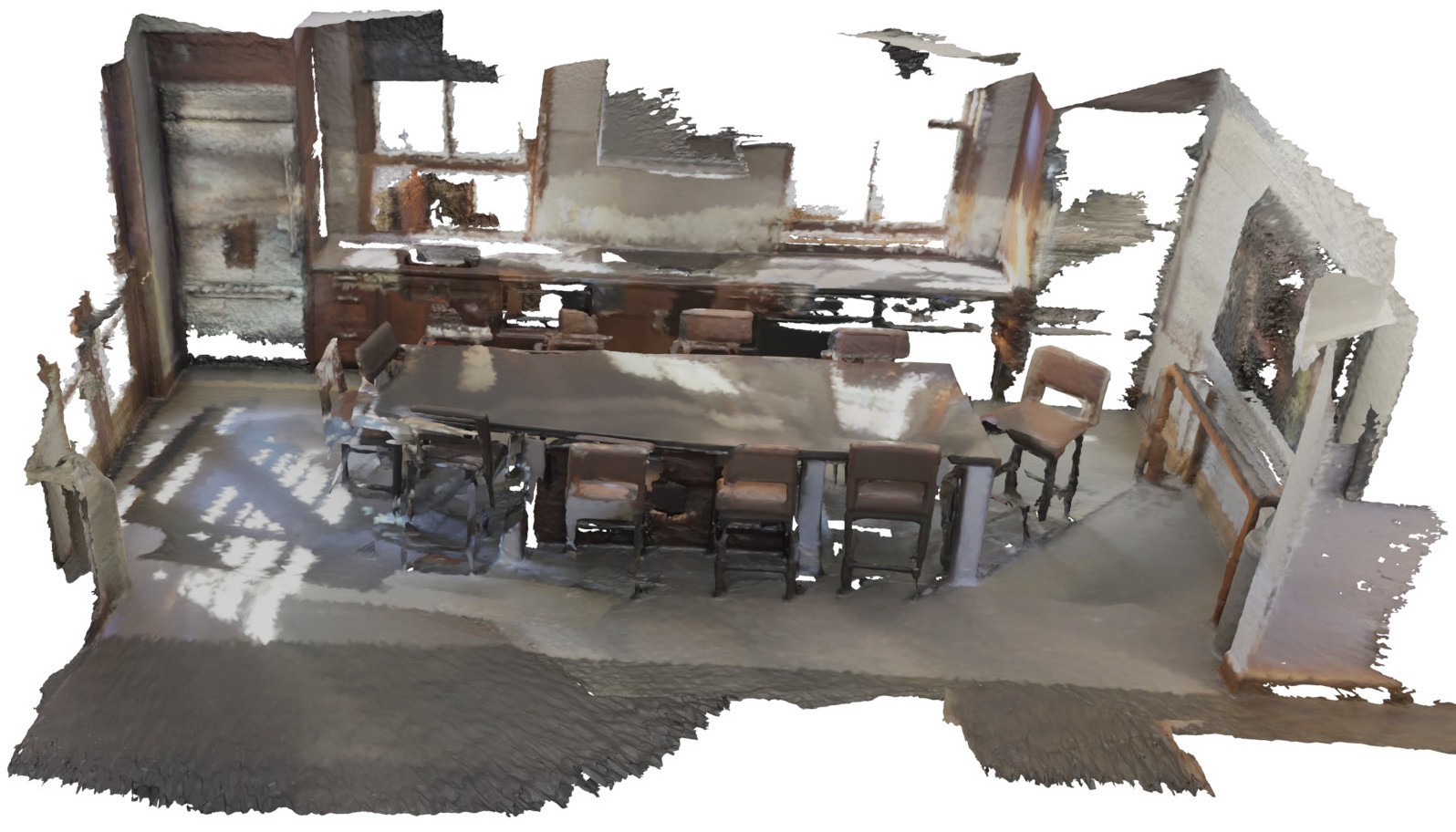}&\includegraphics[width=0.49\textwidth]{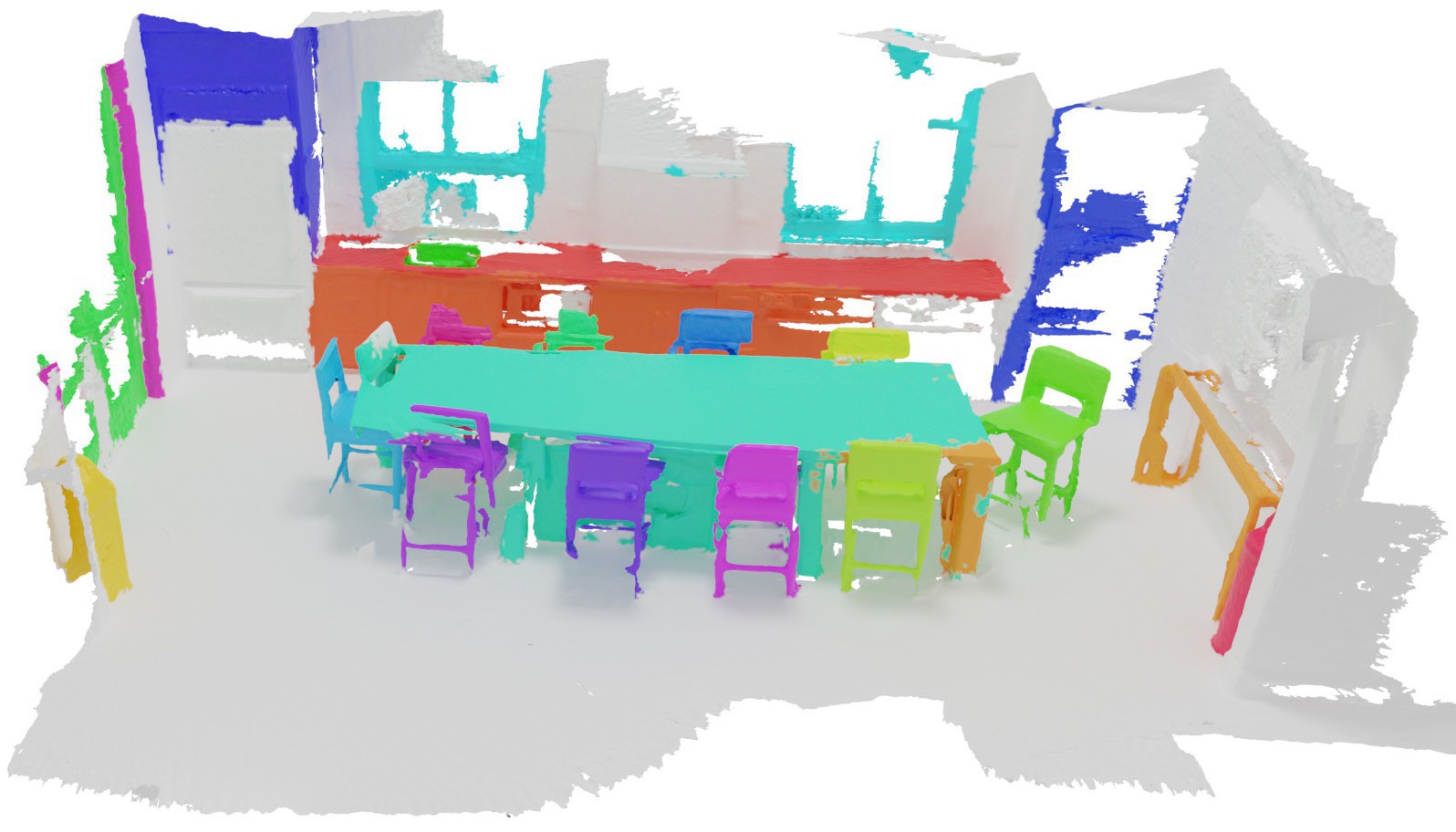} \\
    \textbf{(c)} Instance Heatmap (Windows) & \textbf{(d)} \name{} (ours) with DBSCAN \\
    \includegraphics[width=0.49\textwidth]{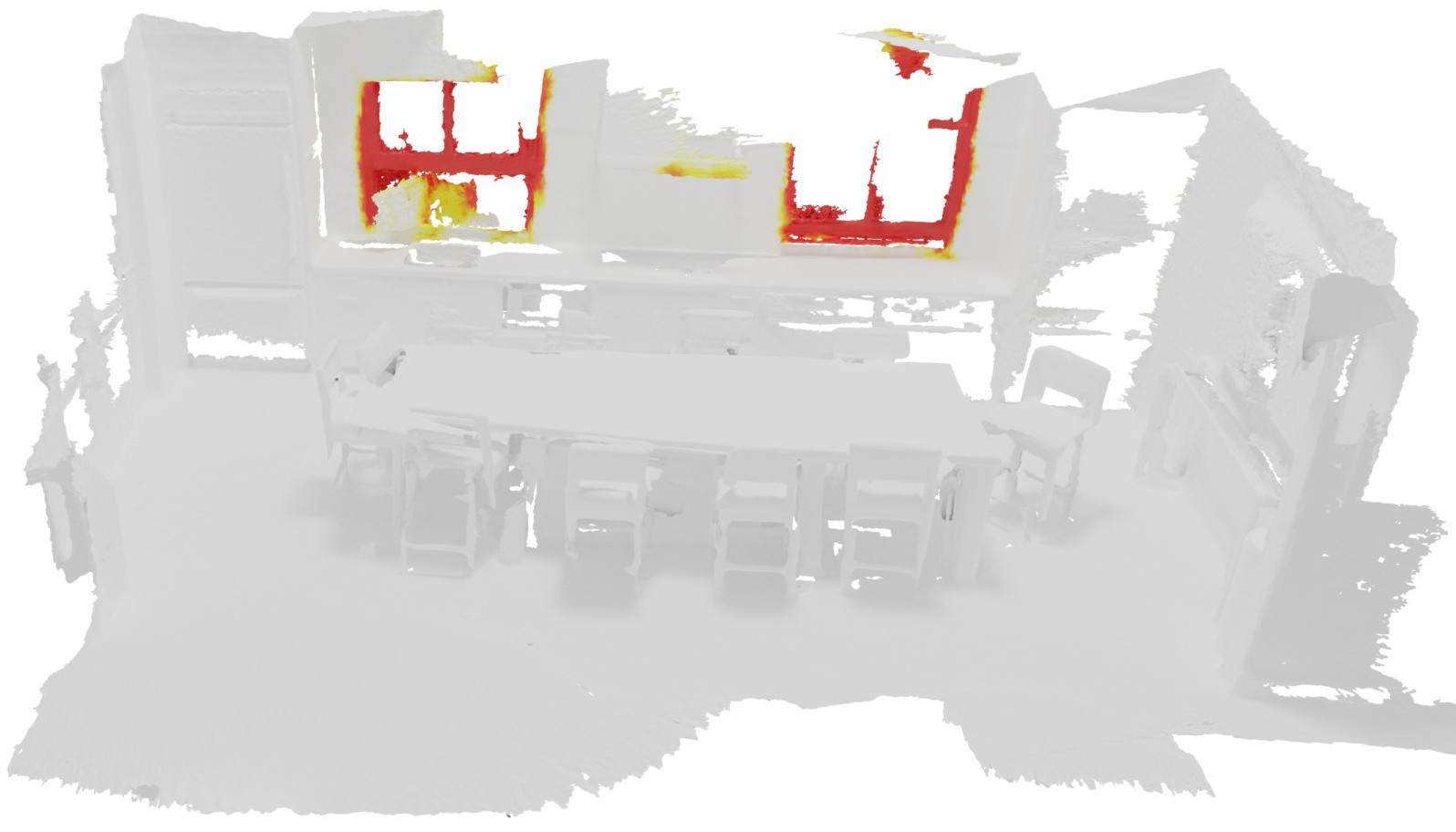}&\includegraphics[width=0.49\textwidth]{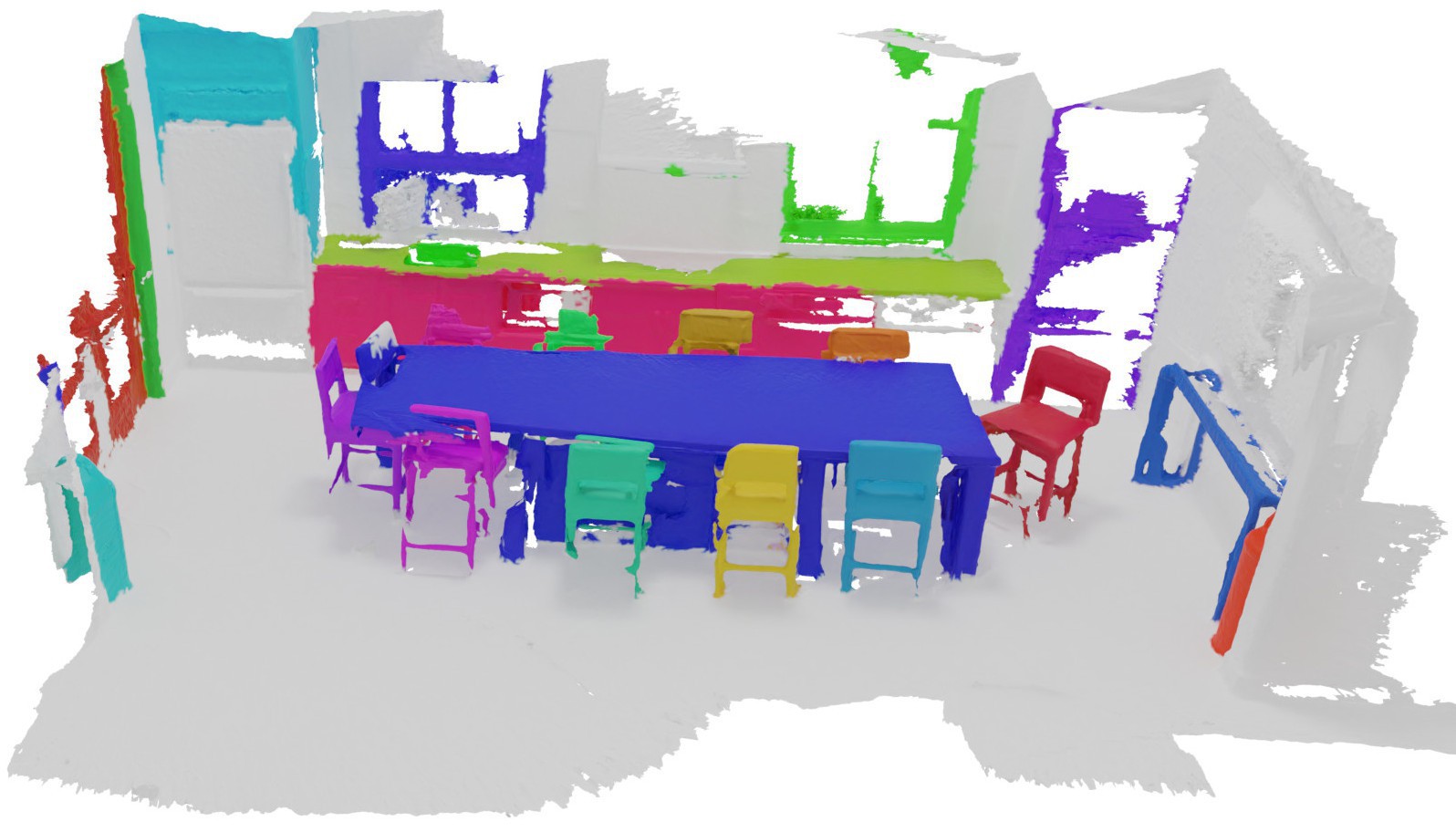} \\
    \midrule
    \textbf{(e)} Center Votes (SoftGroup\,\cite{Vu22CVPR}) & \textbf{(f)} Prediction (SoftGroup\,\cite{Vu22CVPR}) \\
    \includegraphics[width=0.49\textwidth]{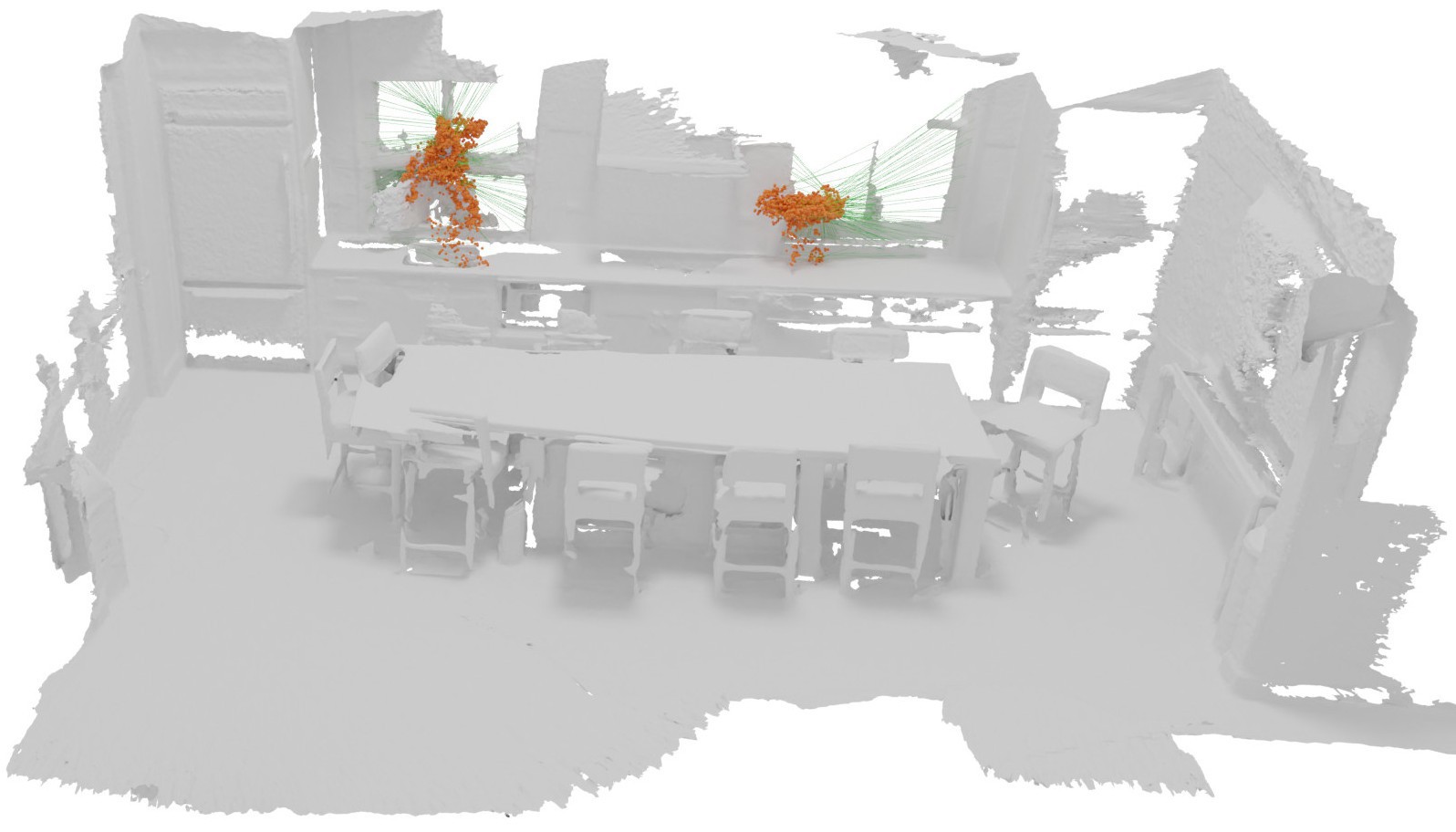}&\includegraphics[width=0.49\textwidth]{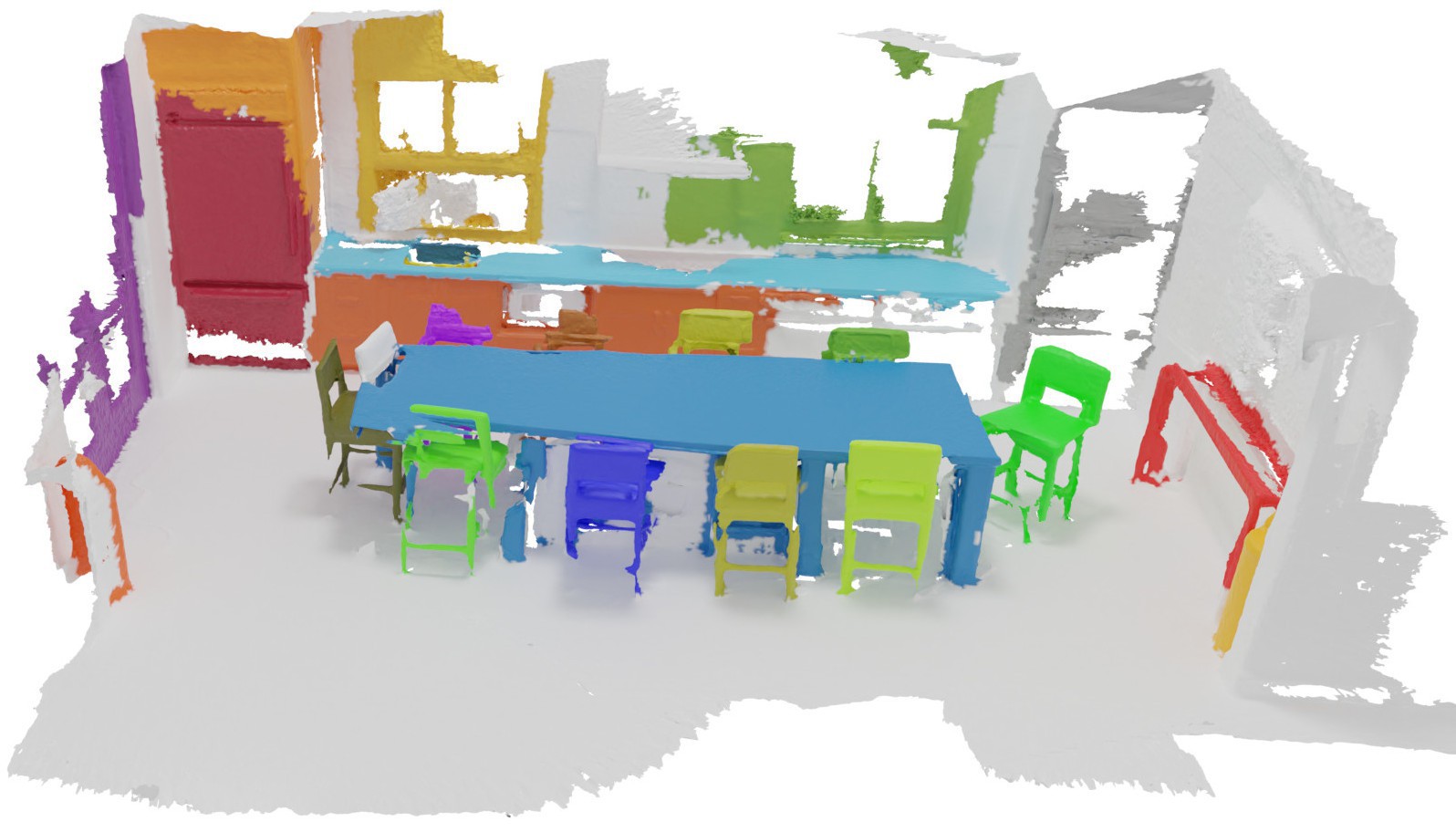} \\
    \end{tabularx}%
    \caption{%
        \textbf{Qualitative Analysis of DBSCAN Postprocessing.}
        \name{} occassionally predicts masks containing two instances of the same class.
        In \textbf{(b)}, two windows are merged into a single instance since their underlying point cloud features result in a high response when convolved with the instance query (\cf heatmap in \textbf{(c)}).
        In \textbf{(d)}, we apply DBSCAN as a postprocessing routine to split erroneously merged instances based on spatial contiguity.
        We do not see this effect for voting-based methods as they explicitly encode geometric priors \textbf{(e)-(f)}.
    }
    \label{fig:qualitative_dbscan}
\end{figure*}
In Section 4.3 (main paper), we described one limitation of \name{}.
A few times, we observed that similarly looking objects are merged into a single instance even if they are apart in the input point cloud (\cf \reffig{qualitative_dbscan}(b)--(c)).
We trace this back to \name{}'s possibility to attend to the full point cloud combined with instances which show similar semantics and geometry.
As a solution, we propose to apply DBSCAN\,\cite{Ester96KDD} on the output instance masks produced by \name{}.
For each of the $K$ instance masks individually, DBSCAN {yields} spatially contiguous clusters (\cf \reffig{qualitative_dbscan}(d)).
We treat these dense clusters as new instance masks.
We update the confidence score for each newly created instance by applying Equation (7, main paper).
\begin{table}
	\centering

	\caption{
	\textbf{Ablation on DBSCAN postprocessing.}
	To split wrongly merged instances, we employ DBSCAN as an optional postprocessing routine.
	We report best scores around a minimal distance $\epsilon$=$0.9$ (ScanNet) and $\epsilon$=$0.6$ (S3DIS-A5).
	}
 
	\setlength{\tabcolsep}{1.5px}
	\begin{tabularx}{\linewidth}{YcYYYcYYY} 
	    \toprule
        && \multicolumn{3}{c}{ScanNet Validation ({2\,cm})} && \multicolumn{3}{c}{S3DIS Area 5 ({2\,cm})} \\
	    \cmidrule{3-5} \cmidrule{7-9}
		$\epsilon$  && AP            & AP$_{50}$ & AP$_{25}$ && AP            & AP$_{50}$ & AP$_{25}$ \\
		\midrule
		-- && 54.3 & 73.0 & 83.4 && 55.7 & 69.8 & 76.1 \\
		\midrule
		0.5 && 54.1 & 72.1 & 82.1 && 57.6 & 71.7 & \textbf{77.2} \\
		0.6 && 54.4 & 72.4 & 82.4 && \textbf{57.8} & \textbf{71.9} & \textbf{77.2} \\
		0.7 && 54.9 & 73.2 & 83.1 && 57.7 & 71.8 & \textbf{77.2} \\
  		0.8 && 55.0 & 73.3 & 83.2 && 57.5 & 71.6 & 77.1 \\
       0.9 && \textbf{55.1} & \textbf{73.7} & \textbf{83.6} && 57.6 & 71.6 & 77.1 \\
       1.0 && 55.0 & 73.5 & 83.5 && 57.5 & 71.5 & \textbf{77.2} \\
       1.1 && 55.0 & 73.6 & 83.6 && 57.5 & 71.4 & \textbf{77.2} \\
		\bottomrule
	\end{tabularx}
	\label{tab:score_thr}
\end{table}
In our hyperparameter ablation study in \reftab{score_thr}, we achieved overall best results when applying DBSCAN with a minimal distance parameter $\epsilon$ of $0.9$ for ScanNet, $0.6$ for S3DIS and $14.0$ for STPLS3D.
Note that we do not consider noise points, \ie, we set the minimal size of a cluster to $1$.

}

\end{document}